\documentclass[acmtog,bookmarksnumbered,unicode,pagebackref=true,breaklinks=true,colorlinks,bookmarks=false]{acmart}

\usepackage{times}
\usepackage{epsfig}
\usepackage{graphicx}
\usepackage{amsmath}
\usepackage{soul}
\usepackage{capt-of}
\usepackage{multirow}
\usepackage{adjustbox}
\usepackage[thinlines]{easytable}
\usepackage{gensymb}
\usepackage{booktabs}
\usepackage{subcaption}
\usepackage{mfirstuc}

\def\cvprPaperID{4869} 
\def\confYear{CVPR 2021}

\newcommand{\todo}[1]{}
\newcommand{\IA}[1]{intelligent augmentation}
\newcommand{\AD}[1]{adversarial independence}
\newcommand{\HI}[1]{RENATA}

\setcopyright{none}
\copyrightyear{2020}
\acmYear{2020}
\acmDOI{}
\acmJournal{PRE}

\begin{document}

\author{Neil Joshi}
\affiliation{%
  \institution{Johns Hopkins University Applied Physics Laboratory}
  \streetaddress{11100 Johns Hopkins Road}
  \city{Laurel}
  \state{MD}
  \country{USA}}
\email{neil.joshi@jhuapl.edu}

\author{Phil Burlina}
\affiliation{%
  \institution{Johns Hopkins University Applied Physics Laboratory}
  \streetaddress{11100 Johns Hopkins Road}
  \city{Laurel}
  \state{MD}
  \country{USA}}
\email{pburlin2@jhu.edu}

\title{AI Fairness via Domain  Adaption \\  Application to AMD Detection}

\begin{abstract}
While deep learning (DL) approaches are reaching human-level performance for many tasks, including for diagnostics AI, the focus is now on  challenges possibly affecting DL deployment, including AI privacy, domain generalization, and fairness. This last challenge is addressed in this study.
Here we look at a novel method for ensuring AI fairness with respect to protected or sensitive factors.
This method uses domain adaptation via training set enhancement  to tackle bias-causing training data imbalance. More specifically, it uses  generative models that allow the generation of more synthetic training samples for underrepresented populations. 
This paper applies this method to the use case of detection of age related macular degeneration (AMD). Our experiments show that starting with an originally biased AMD diagnostics model the method has the ability to improve fairness.

\end{abstract}
\maketitle

\section{Introduction}
Deep learning (DL)  has now widely and successfully been applied to tasks ranging from  recognizing faces, detecting people, prescreening diseases or performing lesion segmentation~\cite{pekala2019deep,burlina2019assessment}, and has outperformed past machine learning approaches \cite{burlina2011automatic}. However, new challenges have emerged and are potentially impacting future deployment of AI, including: privacy issues~\cite{shokri2017membership}; challenges that  entail the existence of adversarial machine learning or data poisoning to defeat DL models~\cite{carlini2017adversarial}; the lack of availability of sufficient training data ~\cite{burlina2020low}; or AI fairness which may lead to lack of parity in prediction performance~\cite{burlina2020addressing}, the last concern is our focus here. With regard to AI fairness one important cause of bias is data imbalance which is specifically addressed herein. 

Addressing AI fairness has recently gotten significant attention~\cite{mehrabi2019survey,hardt2016,bolukbasi2016man,kinyanjui2020fairness,quadrianto2019discovering}. Work in addressing AI bias falls  roughly into the following basic taxonomy and the following types of approaches: some methods alter the model, some alter the data, and some perform a form of post-processing via re-calibration.

Regarding approaches that alter the model one path that is now intensely investigated looks at adversarial two player methods that tackle the possible conditional dependence of predictions made by DL models (e.g. predicting AMD) on protected attributes, and use an adversarial technique~\cite{goodfellow2014generative} to promote independence and fairness (see ~\cite{ganin2016domain, beutel2017data} or studies like \cite{wadsworth2018achieving} and \cite{zhang2018mitigating}).
\cite{beutel2017data} in particular uses an  adversarial player that predicts protected factors from the feature embeddings (as does~\cite{wadsworth2018achieving}) with the adversarial network trying to minimize a loss function related to making an accurate attribute prediction, while the main network is both trying to estimate the  labels for the task at hand (for example in our case AMD vs. no AMD) while disallowing the second adversarial network from guessing correctly the protected factors~\cite{zhang2018mitigating}. A related approach is used in~\cite{song2019learning}  via employing a loss function that uses information theoretic measures. \cite{quadrianto2019discovering} uses a method that employs a related adversarial  approach but makes changes to the image (instead of making changes to the embedding representation) and applies this to masking of visual features in faces. 

One possible drawback of this first type of approaches is that a) adversarial two player systems can yield  challenges for training the networks and b) these methods do not address inherent issues of bias emanating from data imbalance. This motivates the type of approach used here which instead uses training data augmentation and domain adaptation. 

To address this important cause of bias, our approach proceeds via synthesizing more data for underrepresented populations and does this via a generative method which has the unique ability to perform fine control of protected attributes, performing a sort of domain adaptation. This strategy has yet to be explored to a large extent compared to the adversarial approach to bias mitigation, and is a promising area of research that is focused on specific aspects of generative models, which are briefly reviewed next. 

Generative models have the ability to generate new data \cite{karras2019style} and therefore the potential of addressing the data imbalance challenges in bias. Generative methods --broadly speaking-- learn to sample from the underlying training data distribution so as to generate new samples who are statistically indistinguishable from the training distribution. Those models broadly fall into several categories that encompass: generative adversarial networks (GANs)~\cite{karras2019style,grover2019bias}, autoencoders, variational autoencoders (VAEs)~\cite{kingma2013auto, louizos2015variational,zhao2017infovae},  generative autoregressive models, and invertible flow-based latent vector models.

Generative methods have evolved  and have culminated recently in approaches leading to  GANs that achieve realism on high resolution images (a former limitation of GANs). This includes  approaches such as BigGAN~\cite{brock2018large}, which relies on SAGAN (self-attention GANs).  ~\cite{brock2018large} exploits larger batch sizes which appear to improve performance. That study noted also that larger networks has a comparable positive effect. Another positive effect results from usage of the truncation method which consists of, for the generator, sampling from a standard normal distribution in training, while sampling instead from a truncated normal distribution at inference/generation time, where samples above a certain threshold are re-sampled. Truncating with a lower threshold allows to  trade-off between higher fidelity and lower diversity. While this model has resulted in many improvements to yield high-resolution image synthesis, it did not aim for alteration of the image akin to style transfer.

As an alternative,  StyleGAN~\cite{karras2019style} used a multi-scale design that was successful at addressing the generation of high resolution images (512x512 or 1024x1024), and  allowed for stylistic image alterations. In that approach a style vector W is employed to affect factors of the image (at the low scale, coarse factors like skin tone, and at higher scales, fine factors like hair). As such this method is able to perform  stylistic  mixing.

However all these methods have limitations for addressing the problem of data imbalance we consider here because -- while they generate realistic images -- these generative actions are  essentially uncontrolled alterations to the image. Even when the method allows style mixing~\cite{karras2019style} this mixing does not allow specific control of a given factor  in the latent space $Z$ or style space $W$, without alteration of other concomitant (entangled) factors.  In sum, those methods, and derivative approaches including  the method in~\cite{burlina2019assessment}
 fail to generate images specifically for a missing or under represented protected factor (eg retinal images for a specific subpopulation, say African Americans with AMD lesions). 
 
 This motivates the search for other methods that allow fine control of individual semantic images factors (e.g. images of dark skin individuals with Lyme disease~\cite{burlina2020ai}). 
While work has emerged that starts to address that problem ~\cite{paul2020unsupervised}, the challenge is that when control of attributes is achieved it may come at the cost of entanglement among latent factors that controls those attributes. While disentanglement was linked to fairness empirically in~\cite{Locatello2019OnTF}, methods that achieve this disentanglement are yet to be fully fleshed out.
These limitations motivate our  approach here, which uses  latent space manipulation  to control specific attributes  while keeping other attributes unchanged.

In summary:  fairness is achieved here via generation of more training data and  this generation is done in a way that  finely controls attributes; This method is able to perform debiasing when applied to the specific use case of age related macular degeneration studied here.
 
\section{Methods}

The general strategy is as follows: our method performs generation of additional  data with fine control of attributes. It does this for attribute conditions and populations less or not represented originally in the training dataset to aim to reduce dataset imbalance. For the specific use case of AMD the goal is to generate more fundus images for underrepresented populations (e.g. dark skin individuals with AMD). This explains why we desire to generate data with fine control of attributes, e.g. change markers for ethnicity or race as well as image markers for the disease, while leaving the rest of the image and image markers unaffected.

These  images are generated as follows: starting with  healthy African American individuals our goal is to generate new images that include lesions but keep other image characteristics unchanged such as vasculature and possibly other image markers like those described in~\cite{poplin2018prediction}. 
The approach uses StyleGAN as a basis. StyleGAN includes a fully connected network that takes a latent vector $Z$ and transforms it into a 512-length intermediate latent vector $W =  \{w_i\}$ for all scales $i$  that controls a multiscale generation process via an adaptive instance normalization (AdaIN) operation and allows stylistic mixing of images (hence the so called 'style' vector). This style vector $W$ was originally used in StyleGAN to affect some factors of the image at the low scale, coarse factors like skin tone, and at the higher scale, fine factors like hair, by doing essentially style mixing. Direct style mixing cannot be used for our purposes which require controlled data generation since style mixing -- along with some useful characteristics changes added to the image (e.g. image lesions) -- may impact other characteristics and other image markers and changes that are undesirable (e.g. changes in the vasculature). 

Our process works as follows: it first trains StyleGAN on the available set of training data, which is biased. It then uses a generated starter image sample in latent space that corresponds to a retina of an African American individual; we denote this by the tuple (image. latent vector) ($I$, $\{w_i\}$), for all scales i in StyleGAN. Then it moves this sample along a trajectory in  latent space that takes its initial position to a location in that space that entails it having more disease lesions (AMD), turning the initial $\{w_i\}$ into $\{w_i^\prime\}$. This new latent space vector $\{w_i^\prime\}$ sample is then turned into an actual image $I^{\prime}$ via the generative model. This process is repeated for as many samples that are needed to reestablish balance in the dataset. Once a balanced dataset is generated a new AMD diagnostics model is then trained with the augmented data and performance is compared to the originally biased model.

To achieve this trajectory in latent space a gradient descent direction of motion is used. This gradient descent is used to maximize a loss function that is obtained as follows: A simple classifier (a neural network)  is trained on the original images to distinguish AMD vs no AMD affected retinas. This classifier is then used to classify synthetic samples of retinas and those samples representations in latent space are used to train a new classifier (AMD vs. no AMD) that now operates in latent space. This second classifier's output loss function, evaluated for a desired label that corresponds to having AMD, is then used as loss function in  the gradient decent process. The net effect of the gradient decent is to 'impart' the disease (AMD) to the original image with minimal alteration of the rest of the image. Hence a fundus image from a African American individual that is likely healthy with regard to AMD is transformed into a realistic fundus retinal image of an African American individual that is AMD-affected.

\section{Experiments}

To measure the effect of the debiasing on an originally biased classifier for AMD we look at several metrics principally: accuracy, sensitivity, specificity, F1, kappa score, PPV and NPV, along with average precision and ROC AUC.

\begin{figure}[t]
\label{example}
\centering
\caption{Examples of African American fundus images, starting with healthy (left), transformed to affected (right), through our domain adaptation approach.}
\includegraphics[width=0.7\linewidth]{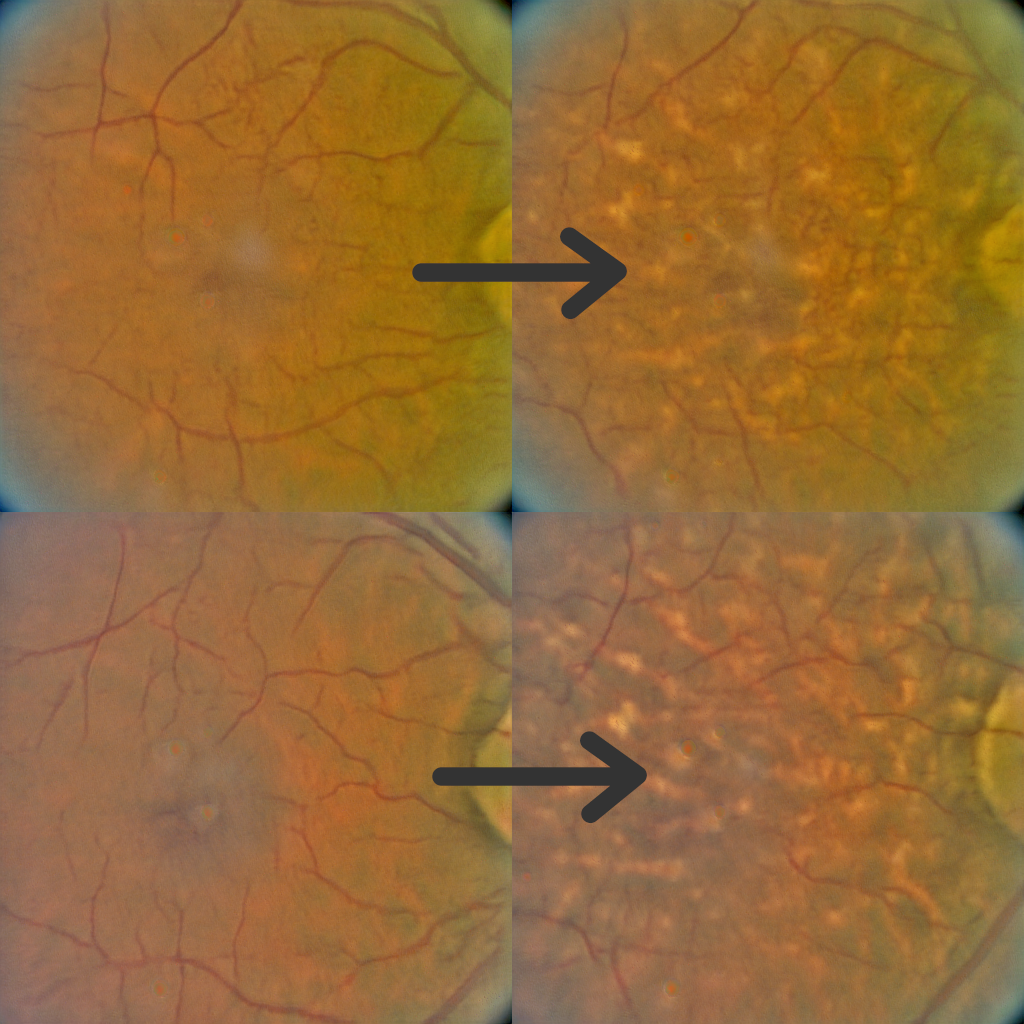}
\end{figure}

\begin{table}[thb!]
\scriptsize
    \begin{center}
        \begin{tabular}{l||c|c}
        \toprule

            Metrics & Baseline  & New Method   \\
   & Method  &  Domain Adaptation \\
        \hline
        \midrule

Accuracy
&
71.75 (5.03)
&
82.79 (4.22)
\\
Sensitivity
&
59.09 (7.77)
&
74.03 (6.93)
\\
Specificity
&
84.42 (5.73)
&
91.56 (4.39)
\\
PPV
&
79.13 (7.43)
&
89.76 (5.27)
\\
NPV
&
67.36 (6.62)
&
77.90 (6.04)
\\
Weighted Kappa
&
0.4351
&
0.6558
\\
F1
&
0.6766
&
0.8114
\\
Average Precision
&
0.8469 (0.0402)
&
0.9284 (0.0288)
\\
ROCAUC
&
0.8049 (0.0443)
&
0.9169 (0.0308)
\\
\midrule
Test Set Subset Analysis:
&
&
\\
\hline
Accuracy (Caucasians)
&
80.52 (6.26)
&
85.71 (5.53)
\\
Accuracy (African Americans)
&
62.99 (7.63)
&
79.87 (6.33)

\\
\midrule
Larger Leftover Set Analysis (614 images):
&
&
\\
\hline
Accuracy (Leftover dataset of African Americans/AMD)
&
42.51 (3.91)
&
54.23 (3.94)
\\

            \hline
        
        \bottomrule
        \end{tabular}
    \end{center}
\caption{Results for debiasing comparing results for the baseline model and the synthetic debiased model using latent space manipulation showing performance improvement as well as debiasing for the proposed approach.}
\label{tbl:AMD_results}
\end{table}

\begin{table}[tbh!]
\scriptsize
    \begin{center}
        \begin{tabular}{l||c|c}
        \toprule

            Test data & Healthy& Affected (AMD) \\

        \hline
        \midrule
            
          Caucasians & 1843 & 1843 \\
            \hline
            
           African Americans & 3686 & 0 \\
            \hline
            
        \bottomrule
        \end{tabular}
    \end{center}
\caption{Characteristic table for training dataset for baseline AMD diagnostic model.}
\label{tbl:characteristic_training}
\end{table}

\begin{table}[tbh!]
\scriptsize
    \begin{center}
        \begin{tabular}{l||c|c}
        \toprule

            Test data & Healthy& Affected (AMD) \\

        \hline
        \midrule
            
          Caucasians & 77 & 77 \\
            \hline
            
           African Americans & 77 & 77 \\
            \hline
            
        \bottomrule
        \end{tabular}
    \end{center}
\caption{Characteristic table for testing dataset. This dataset is used both for the baseline as well as the new (domain adapted) diagnotics model.}
\label{tbl:characteristic_testing}
\end{table}

We use a retinal disease use case focused on AMD that is based on using the AREDS dataset. AREDS is a dataset~\cite{burlina2011automatic,burlina2019assessment} made available by request to the NIH. It contains individuals' retinas with labels corresponding to a 4-scale AMD severity.  It also includes information on ethnicity. Individuals self declared as Caucasian or African American are used in this study. 

We built a model that predicts no AMD (severity 1 and 2) vs. referable AMD (severity 3 and 4). We employ  a use case of  data imbalance where the training partition was such that no data was made available to the baseline model for African American individuals that were affected by AMD, all other types of data were made available. The training dataset is described in the characteristic table~\ref{tbl:characteristic_training} for training. Note that an equal number of healthy and diseased retinal fundi were used in training so as not to incur artificial bias in the disease prediction overall. Only a reduced number of images were left over for testing considering we had to equally balance those partitions resulting in Table~\ref{tbl:characteristic_testing}. A left over dataset of retinal images from African American individuals with AMD  was used for additional testing (about 614 images). 
Results are shown in~\ref{tbl:AMD_results} comparing the original baseline AMD diagnostics model and the model. It is apparent looking at these results that the baseline model trained on the originally unbalanced dataset exhibits bias (Accuracy on Caucasian is 80.52\% (6.26\%) vs  African American individuals, 62.99\% (7.63\%)) while the new model trained on the augmented dataset has accuracies of 85.71\% (5.53\%) for Caucasian individuals vs 79.87\% (6.33\%) for African American individuals.

\section{Discussion}
\label{sec:Discussion}

The proposed method that generates more data via latent space manipulation and is able to address domain adaptation shows performance  improvements in regard to most metrics (e.g. the baseline system has a ROC AUC of 0.8049 (0.0443) vs an ROC AUC of 0.9169 (0.0308) for the new diagnotic model using augmented data via latent space manipulation).  

What is noticeable however  is that the new diagnostic model is able to improve overall performance as well as reduce the performance gap between populations. 

Looking at the 95\% confidence interval allows us to conclude positively that the performance improvement in metrics such as accuracy and ROC AUC between baseline and new system. Looking  at the overall system performance also indicates similar improvements when considering the other performance metrics.

Similarly, looking at accuracy and noting the difference between the accuracy for Caucasian and African American individuals shows significance in terms of the improvement made, from baseline model to new model;  it shows that there was definitely a bias in the original model;  it also shows that this bias (gap in accuracy between performance of African American individuals and Caucasian individuals)  became within  confidence interval for the new diagnostic model, showing that the new model was more fair. The metric measuring accuracy on a leftover dataset of African American individuals with AMD also demonstrated similar improvement in performance in the new model.

Future work will entail the development of novel metrics to better understand the tradeoffs between accuracy and fairness which are not discussed here.

\section{Conclusion}

We study a method for debiasing that generates more data for underrepresented populations and addresses domain adaptation as a means of tackling bias.  We show that we are able to use this method to address AI fairness in a use case that entails developing a DL based detection system for  age related macular degeneration. The experiments show that this approach is indeed able to improve fairness.


{\small
}

\bibliographystyle{ieee_fullname}

\end{document}